\documentclass[AMA,Times1COL]{WileyNJDv5} 

\articletype{Article Type}%

\received{Date Month Year}
\revised{Date Month Year}
\accepted{Date Month Year}
\journal{International Journal}
\volume{00}
\copyyear{2024}
\startpage{1}

\usepackage{graphicx}      
\usepackage{natbib}        
\usepackage{amsmath}
\usepackage{amsfonts}
\usepackage{amssymb}
\usepackage{upgreek}
\usepackage{nicefrac}
\usepackage{mathrsfs}
\usepackage{xurl}
\usepackage{subcaption}
\newcommand{\Sabs}[1]{\left\lceil #1 \right\rfloor}

\newcommand{\norm}[1]{\left|\left| #1 \right|\right|}

\newcommand{\sign}{\textup{sign}}

\begin{document}

\title{AI-Driven approach for sustainable extraction of earth's subsurface renewable energy while minimizing seismic activity}

\author[1]{Diego Guti\'errez-Oribio}
\author[1]{Alexandros Stathas}
\author[1]{Ioannis Stefanou}

\authormark{Guti\'errez-Oribio \textsc{et al.}}
\titlemark{AI-Driven Solution for Sustainable Extraction of Earth's Subsurface Renewable Energy while Minimizing Seismic Activity}

\address[1]{\orgname{Nantes Universite, Ecole Centrale Nantes, CNRS, GeM, UMR 6183, F-44000}, \orgaddress{\state{Nantes}, \country{France}} \email{diego.gutierrez-oribio@ec-nantes.fr, alexandros.stathas@ec-nantes.fr, ioannis.stefanou@ec-nantes.fr}}

\corres{Corresponding author Diego Guti\'errez-Oribio, Nantes Universite, Ecole Centrale Nantes, CNRS, GeM, UMR 6183, F-44000, Nantes, France. \email{diego.gutierrez-oribio@ec-nantes.fr}}

\fundingInfo{European Research Council's (ERC) support under the European Union’s Horizon 2020 research and innovation program (Grant agreement no. 101087771 INJECT) and the Region Pays de la Loire and Nantes M\'etropole under the Connect Talent programme (CEEV: Controlling Extreme EVents - Blast: Blas LoAds on STructures).}


\abstract[Abstract]{
 Deep Geothermal Energy, Carbon Capture and Storage, and Hydrogen Storage hold considerable promise for meeting the energy sector's large-scale requirements and reducing CO$_2$ emissions. However, the injection of fluids into the Earth's crust, essential for these activities, can induce or trigger earthquakes. In this paper, we highlight a new approach based on Reinforcement Learning for the control of human-induced seismicity in the highly complex environment of an underground reservoir. This complex system poses significant challenges in the control design due to parameter uncertainties and unmodeled dynamics. We show that the reinforcement learning algorithm can interact efficiently with a robust controller, by choosing the controller parameters in real-time, reducing human-induced seismicity and allowing the consideration of further production objectives, \textit{e.g.}, minimal control power. Simulations are presented for a simplified underground reservoir under various energy demand scenarios, demonstrating the reliability and effectiveness of the proposed control-reinforcement learning approach.
 
}

\keywords{Energy geomechanics, Earthquake prevention, Reinforcement Learning, Robust control}

\jnlcitation{\cname{%
\author{Guti\'errez-Oribio D.},
\author{Stathas A.}, and
\author{Stefanou I.}}.
\ctitle{AI-Driven Solution for Sustainable Extraction of Earth's Subsurface Renewable Energy while Minimizing Seismic Activity.} \cjournal{\it Journal.} \cvol{ }.}

\maketitle

\renewcommand\thefootnote{}
\footnotetext{\textbf{Abbreviations:} RL, Reinforcement Learning; ML, Machine Learning; SR, Seismicity Rate; DDPG, Deep Deterministic Policy Gradient.}

\renewcommand\thefootnote{\fnsymbol{footnote}}
\setcounter{footnote}{1}


\section{INTRODUCTION}
Recently, the industrial world's growing energy demands with the need to slow CO$_2$ emissions that accelerate climate change have motivated scientists and engineers towards new technologies, including Deep Geothermal Energy, Carbon Capture and Storage, and Hydrogen Storage, \cite{b:IPCC2023}. These promising new technologies involve the process of injection of fluids in underground reservoirs, which has the potential to induce earthquakes (\textit{i.e.}, human-induced seismicity, see \cite{b:10.1785/0220170112, b:Rubinstein-Mahani-2015, b:10.1002/2016RG000542}). Indeed, human-induced seismicity, has already prompted the closure of several geothermal plans globally \cite{b:Maheux_FBleu, b:Stey_LeMonde,b:Sun_Hank, b:Zastrow-2019,b:10.1785/gssrl.80.5.784, b:Glanz_NYT}.

In the framework of optimization theory, the industrial objective can be stated as controlling the fluid circulation to minimize human-induced seismicity, while sustaining energy production. This problem involves a system (an underground reservoir) that is highly complex with parameters and dynamics that are not (and can not be) entirely known.

These highly ambitious objectives can be met using Machine Learning (ML) techniques and, in particular, Reinforcement Learning (RL). 
RL focuses on the development of software agents that are capable of making optimal decisions in dynamic and uncertain environments.  The advent of RL can be traced back to the \textit{Dynamic programming} optimisation methods established in \cite{bellman1954theory,bertsekas2012dynamic}. It is a powerful framework that enables machines to learn from their interactions with the environment, rather than relying on explicit instructions or labeled datasets. In RL, an agent learns through a trial-and-error process, where it takes actions in an environment, receives feedback in the form of rewards or penalties, and adjusts its behaviour to maximize the cumulative reward over time (see \cite{geron2022hands,sutton2018reinforcement} for more details). Due to its advantages, RL has been used to optimize the performance of complex systems based on a given reward.

In \cite{b:Timos_Stefanou_2021}, a state-of-the-art asynchronous actor-critic network (A3C) has been implemented for controlling earthquake-like instabilities in a simplified earthquake model (the spring-slider). This model-free approach allows the RL algorithm to learn how to control the system's response by autonomously adjusting the uniform pressure applied on the spring-slider, without requiring any prior knowledge of the environment dynamics.

Nevertheless, in a large reservoir system, the space of the unknown states is huge, since the spatial distribution of the fluid pressure is also taken into account (see ``curse of dimensionality" in \cite{bellman1966dynamic}), leading to important difficulties (\textit{e.g.}, catastrophic forgetting and oscillations, see \cite{geron2022hands,ratcliff1990connectionist}) in completing training when a standard reinforcement learning approach is used.

Inspired by \cite{b:Timos_Stefanou_2021}, we apply RL in the more involved underground reservoir model, where diffusion of the injection fluid and the seismicity rate (SR) of the region are accounted for. To allow the agent to learn in this higher dimensional space, we employ a new approach, in which a robust controller (see \cite{b:Gutierrez-Stefanou-2024}) is implemented to control the fluid injection over the reservoir and the RL will adjust the controller gains automatically depending on a given optimization task. This is known as gain-scheduled reinforcement learning \cite{b:HOSSEINI2020897,b:machines9120319,b:timmerman2024adaptive}.

More specifically, in \cite{b:Gutierrez-Stefanou-2024}, the authors have provided initial insights into controlling induced seismicity in underground reservoirs using robust control techniques, while considering fluid circulation constraints linked to energy production. These controllers are adept at addressing model uncertainties and disturbances within the system. However, their effectiveness hinges on accurate knowledge of the bounds associated with these uncertainties and disturbances, which can be challenging to measure accurately in real underground reservoirs.

This combination of RL with robust control theory allows for the introduction of further objectives in the reward function of the problem. The RL algorithm provides a suitable selection of the controller parameters to meet such goals, \textit{i.e.}, minimizing the SR in a given region while meeting the energy demands and minimizing the control power of the wells.

The paper's structure is the following: In Section \ref{sec:problem}, the seismicity rate (SR) model is introduced, and the problem statement of the work is outlined, illustrating how the SR increases with fluid injections. The combined control-RL strategy for minimizing induced seismicity is presented in Sections \ref{sec:control} and \ref{sec:rl}. To demonstrate the effectiveness of the proposed approach, simulations are conducted in Section \ref{sec:sim}, considering various scenarios of intermittent energy demand and production constraints. Finally, Section \ref{sec:conclusions} provides concluding remarks, summarizing the key findings of the study.

The following notation will be used throughout the text: We denote by $\norm{\cdot}$ the euclidean norm of the $n$-dimensional Euclidean space, $\mathbb{R}^n$. For $y_e \in [C^0(T)]^m$, the function $\lceil y_e \rfloor^{\gamma}:=|y_e|^{\gamma}\sign(y_e)$ is defined for any $\gamma\in \mathbb{R}_{\geq 0}$. For $y_e \in [C^0(T)]^m$, the functions $\lceil y_e \rfloor^{\gamma}$ and $|y_e|^{\gamma}$ will be applied element-wise.

We denote by $\overline{V}\subset \mathbb{R}^3$ the compact domain that contains $V$ an open subset in $\mathbb{R}^3$ of positive measure and $S=\partial V\in C^{0,1}$ its boundary, which is assumed to be Lipschitz continuous. We also define $T=[0,+\infty)$ as the open time domain starting at 0. Consider the scalar functions $u(x,t)$ that belong to the Sobolev space, $\mathcal{W}=C^0(T;H^1(V))$, such that:
\begin{align*}
    \mathcal{W} 
    =\Big\{& u \mid u(x,\cdot), \nabla u(x,\cdot) \in {\cal L}^2(V), \quad \sup_{t\in T}\norm{u}_{H^1(V)}<+\infty,\; \sup_{t\in T}\norm{u_t}_{H^1(V)}<+\infty \Big\},
\end{align*}
where $x \in \mathbb{R}^3$, $x=[x_1,x_2,x_3]^T$, denotes the spatial variable belonging to $V$, $t \in T$ represents the time variable, and $\norm{u}_{H^1(V)}=\norm{u}_{{\cal L}^2(V)}+ \norm{\nabla{u}}_{{\cal L}^2(V)}$, $\norm{\cdot}_{{\cal L}^2(V)}=\left(\int_V \cdot^2dx\right)^{1/2}$. Moreover, we denote by $u_t =\nicefrac{\partial u}{\partial t}$ the derivative w.r.t. time, by $\nabla u=\left[\nicefrac{\partial u}{\partial x_1},\nicefrac{\partial u}{\partial x_2},\nicefrac{\partial u}{\partial x_3}\right]$ the gradient, and by $\nabla^2 u=\nicefrac{\partial^2 u}{\partial x_1^2}+\nicefrac{\partial^2 u}{\partial x_2^2}+\nicefrac{\partial^2 u}{\partial x_3^2}$ the Laplacian. 

We define the Delta sequence, $\delta(x)$, as a sequence that converges to the Delta (Dirac’s) distribution defined as $\int_{V^*} \phi(x) \delta(x-x^*)\, dV=\phi(x^*)$, $\forall$ $x^* \in V$, $V^* \subset V$, on an arbitrary test function $\phi(x) \in H^1(V)$.

\section{PROBLEM DESCRIPTION AND STATEMENT}
\label{sec:problem}

\begin{figure*}[ht!]
  \centering 
  \includegraphics[width=13cm,keepaspectratio]{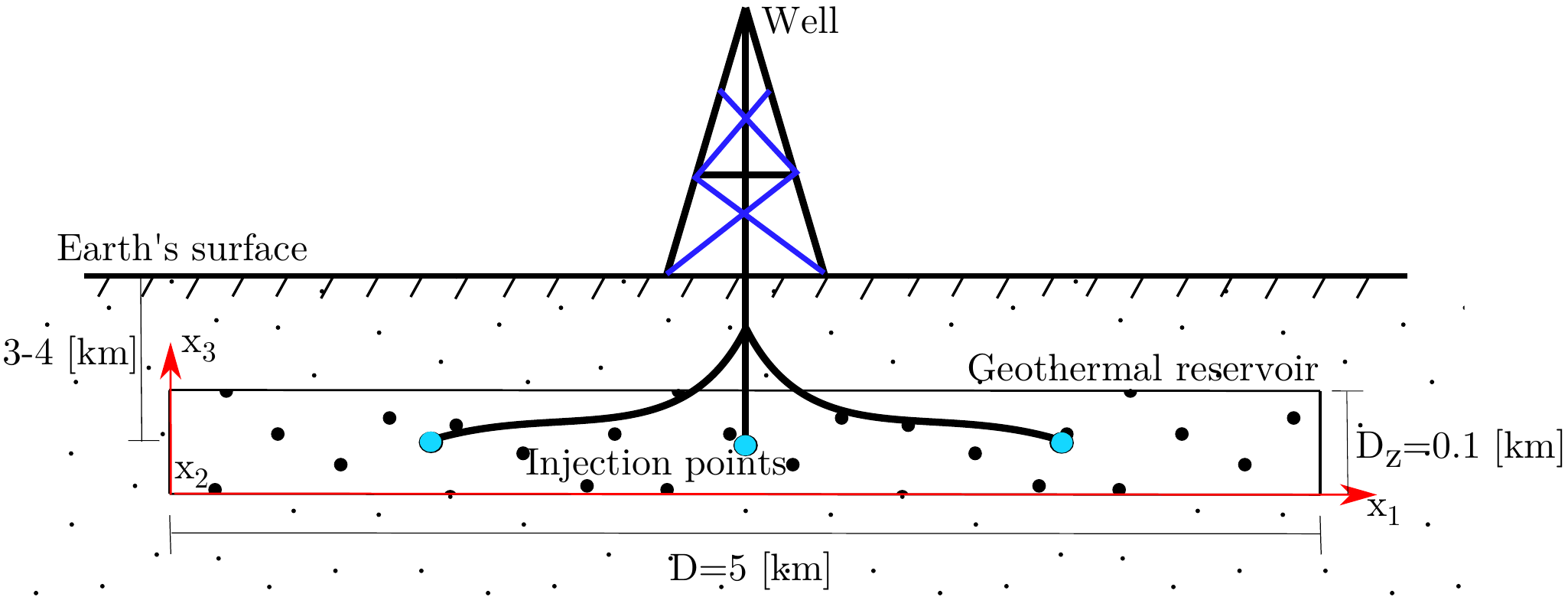}
  \caption{Underground reservoir diagram (see also \cite{b:Gutierrez-Stefanou-2024}).}
  \label{fig:reservoir}
\end{figure*}

Consider a simplified underground reservoir located approximately $4$ [km] below the Earth's surface, as illustrated in Figure \ref{fig:reservoir}. The reservoir comprises porous rock, facilitating the circulation of fluids through its pores and cracks. In this example, the reservoir has a thickness of approximately $100$ [m] and horizontally covers a square surface with dimensions of approximately $5$ [km] by $5$ [km]. Wells are utilized for injecting and/or extracting fluids (e.g., water) at various injection points within the reservoir, as depicted in Figure \ref{fig:reservoir}. For simplicity, the term "injection of fluids" will encompass both injection and extraction operations within the reservoir.

The process of pumping fluids into the reservoir at depth induces fluid circulation, which leads to the deformation of the porous rock hosting the reservoir. The hydro-mechanical behaviour of the reservoir resulting from fluid injections at depth can be described using Biot's theory \cite{b:Biot-1941}. According to this theory, the diffusion of fluid and the deformation of the porous rock are dynamically coupled processes. However, if the injection rates are sufficiently slow compared to the system's characteristic timescales attributable to inertia, and if the volumetric strain rate of the host porous rock is negligible, then the diffusion of fluid within the host rock due to fluid injections can be effectively described by the following diffusion equation \cite{Zienkiewicz1980}
\begin{equation}
\begin{split}
  u_{t}(x,t) &= c_{hy} \nabla^2 u(x,t)+ \frac{1}{\beta}\Big\langle \bar{\mathcal{B}}_c(x), \bar{Q}_c(t)\Big\rangle ,\\
  u(x,t) &= 0 \quad \forall \quad x \in S, \quad t\in[0,T]\\
  u(x,0) &= u^0(x) \in {\cal L}^2(V),
\end{split}
\label{eq:diff}
\end{equation}
where $u(x,t)$ represents the evolution of fluid pressure change within the space $\mathcal{W}$ and $u^0(x)$ its initial condition. The parameters, $c_{hy},\beta \;\in \mathbb{R}$ represent the hydraulic diffusivity, and compressibility of the rock-fluid mixture, which are considered constant.

We consider drained boundary conditions at the boundary of the reservoir, \textit{i.e.}, $u=0$ at $\partial V$. Furthermore, we assume point source terms, as the diameter of the wells is negligible compared to the size of the reservoir. In particular, we define this point-wise fluid injection via the product $\langle\cdot,\cdot\rangle$ between $\bar{B}_c(x), \bar{Q}_c(t)$. We define the vector of the control (well) fluxes applied at the injection points $(x_c^1,...,x_c^{m})$ by $\bar{Q}_c(t)=[\bar{Q}_{c_1}(t),...,\bar{Q}_{c_{m}}(t)]^T \in [C^0(T)]^m$ and define the vector of control coefficients by $\bar{\mathcal{B}}_c(x)= [\delta(x-x_c^1),...,\delta(x-x_c^{m})]^T$ (see \cite{b:Gutierrez-Stefanou-2024} for the rigorous statement of the mathematical problem).

It is now well established that injecting fluids into the Earth's crust can lead to the formation of new seismic faults and the reactivation of existing ones, resulting in significant earthquakes\cite{b:Rubinstein-Mahani-2015,b:Keranen-Savage-Abers-Cochran-2013,b:Zastrow-2019}. The underlying physical mechanism behind these human-induced seismic events is associated with changes in stresses within the host rock caused by the injections, which can either intensify loading or reduce friction along existing or newly formed discontinuities (faults). In simpler terms, fluid injections can elevate the SR in a region, meaning that the number of earthquakes occurring within a given time window increases.

In this study, the seismicity rate (SR) is defined region-wise. We will define the normalized SR over $m_c \in \mathbb{N}$ regions, $V_i \subset V$, $i = 1,...,m_c$, of the underground reservoir as follows
\begin{equation}
\begin{split}
  \dot{R}_i &= \frac{f}{t_a \dot{\tau}_0 V_i} R_i \int_{V_i} u_t(x,t) \, dV - \frac{1}{t_a}R_i(R_i-1), \quad i = 1,...,m_c, 
\end{split}
  \label{eq:SR}
\end{equation}
where $R_i \in C^0(T)$ represents the average normalized SR over a region $V_i$. $f$ represents a mobilized friction coefficient, $t_a$ represents a characteristic decay time, and $\dot{\bar{\tau}}_0$ denotes the background stress change rate in the region, \textit{i.e.}, the stress change rate due to various natural tectonic processes, and all these parameters are assumed to be constant. The equation presented coincides with that of Segall and Lu \cite{Segall2015,Dieterich1994}, with the distinction that here the SR is defined on a region-wise basis rather than point-wise. This choice offers a more generalized and convenient formulation as we primarily focus on averages over large volumes rather than point-wise measurements of the SR, which can also be singular due to point sources (see also \cite{b:Gutierrez-Stefanou-2024}). 

\begin{table*}
\begin{center}
\caption{Diffusion and seismicity rate nominal system parameters \cite{b://doi.org/10.1029/2019JB019134,b://doi.org/10.1002/2015JB012060}.}
\begin{tabular}{ccc}
\toprule 
\textbf{Parameter} & \textbf{Description} & \textbf{Value and Units} \\ 
\midrule
$c_{hy}$ & Hydraulic diffusivity & $3.6 \times 10^{-4}$ [km$^2$/hr] \\ 
$D_x=D_y=D$ & Reservoir's dimension & $5$ [km] \\ 
$D_z$ & Reservoir's thickness & $0.1$ [km] \\ 
$\beta$ & Mixture compressibility & $1.2 \times 10 ^{-4}$ [1/MPa] \\ 
$f$ & Friction coefficient & $0.5$ [-] \\ 
$\dot{\tau}_0$ & Background stressing rate & $1 \times 10 ^{-6}$ [MPa/hr] \\ 
$t_a$ & Characteristic decay time & $500100$ [hr] \\ 
\bottomrule 
\end{tabular}
\label{tab:param}
\end{center}
\end{table*}

\begin{figure}[ht!]
  \centering 
  \begin{subfigure}[T]{0.45\linewidth}
  \caption{}
  \label{fig:reservoir_no}\includegraphics[width=\linewidth,keepaspectratio]{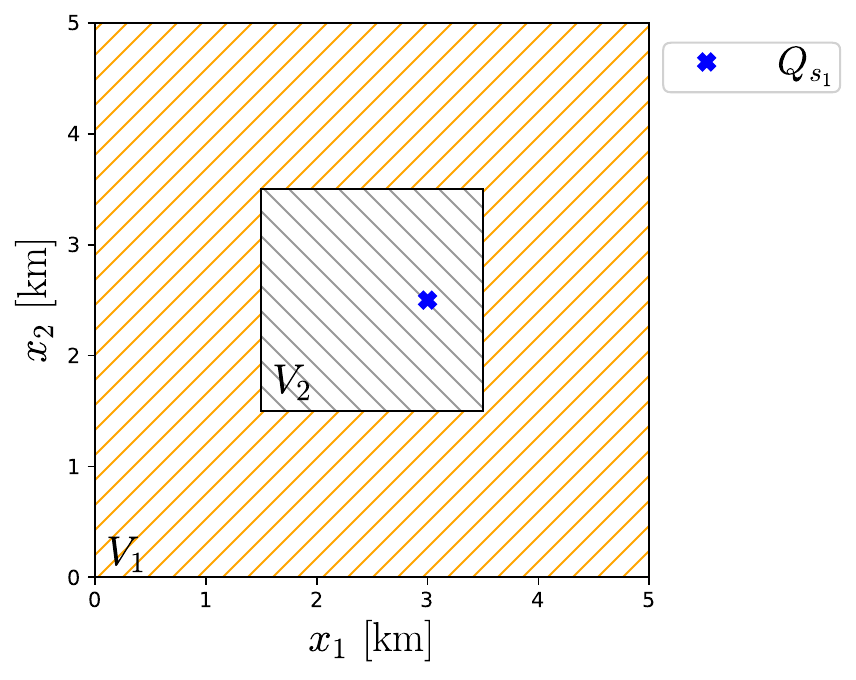}
  \end{subfigure}
  \begin{subfigure}[T]{0.45\linewidth}
  \caption{}
 \label{fig:SR_no} \includegraphics[width=\linewidth,keepaspectratio]{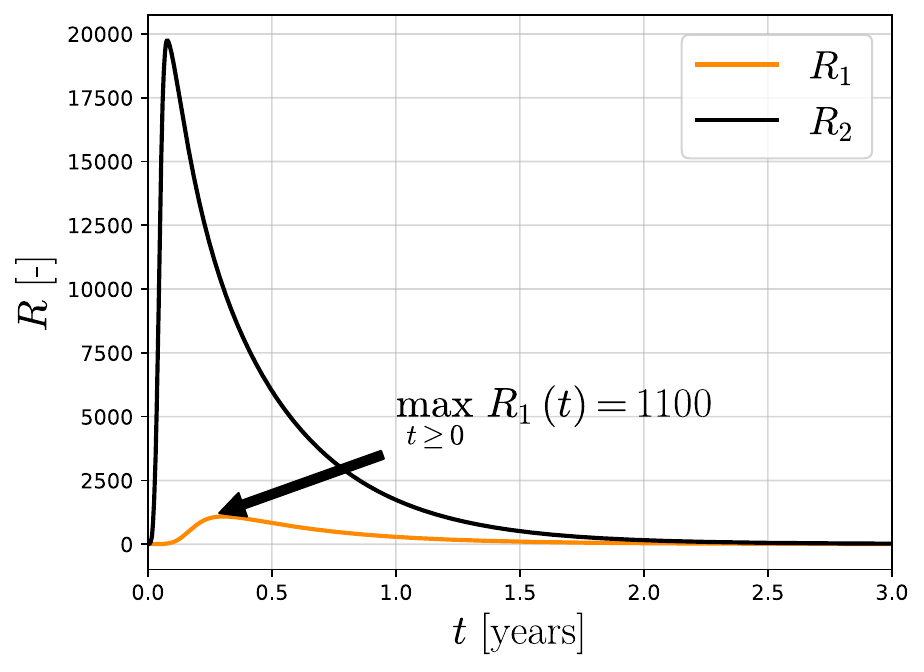}    
  \end{subfigure}
  \caption{a) Regions $V_1$ and $V_2$ and location of the injection well with flux $Q_{s_1}$ inside region $V_2$. b) Seismicity rate in both regions, $V_1,V_2$ with constant injection rate, $Q_{s_1}=15$ [m$^3$/hr]. $1100$ more earthquakes of a given magnitude in a given time window are expected over the outer region of the reservoir due to the constant fluid injection.}
\end{figure}

In the absence of fluid injections, $u_t(x,t)=0$, and therefore, $R_i \rightarrow 1$. In this case, the normalized SR of the region $V_i$ reduces to the natural one. However, if fluids are injected into the reservoir, then $u_t(x,t)>0$, leading to an increase in the SR ($\dot{R}_i>0$) over the region. To illustrate this mechanism, let us consider a static (constant) injection rate of $\bar{Q}_c(t)=Q_{s_1}(t)=15$ [m$^3$/hr] through a single injection well. In this numerical example, we consider the parameters listed in Table \ref{tab:param}, we depth-average Equation \ref{eq:diff} and we integrate the resulting partial differential equation in space and time using a fast Fourier transform method and an explicit Runge-Kutta method of order 3, respectively \cite{bogacki_32_1989,b:Gutierrez-Stefanou-2024}. We then calculate the SR over two distinct regions, one close to the injection point and one in the surroundings (see Figure \ref{fig:reservoir_no} for the location of the regions and the injection point).

We show the SR in both regions as a function of time in Figure \ref{fig:SR_no}. We observe that the maximum SR over $V_1$ is equal to $R_1=1100$. This indicates that over any time period (time window), $1100$ more earthquakes of a given magnitude are expected over region $V_1$ in contrast to the no-injection scenario. The seismicity is even higher near the injection well, as evidenced by $R_2$ in region $V_2$ (see Figure \ref{fig:SR_no}).

\begin{figure}[ht!]
  \centering 
       \includegraphics[width=0.4\textwidth,keepaspectratio]{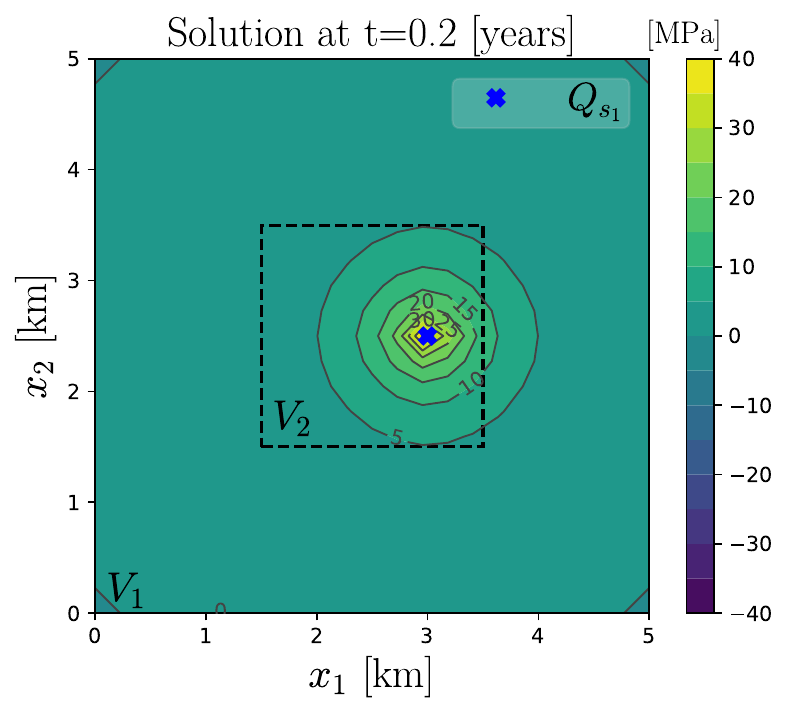}  
       \includegraphics[width=0.4\textwidth,keepaspectratio]{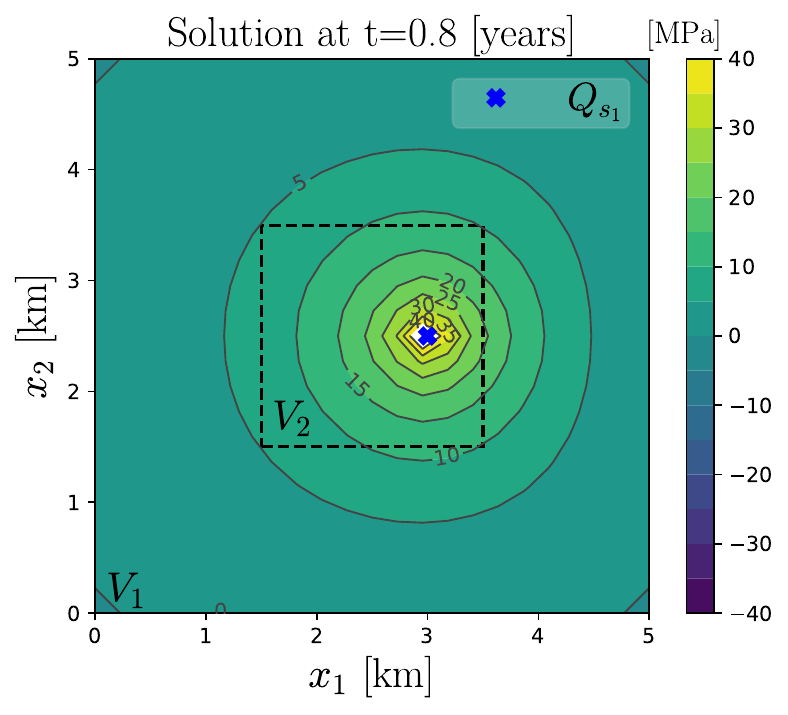}
       \includegraphics[width=0.4\textwidth,keepaspectratio]{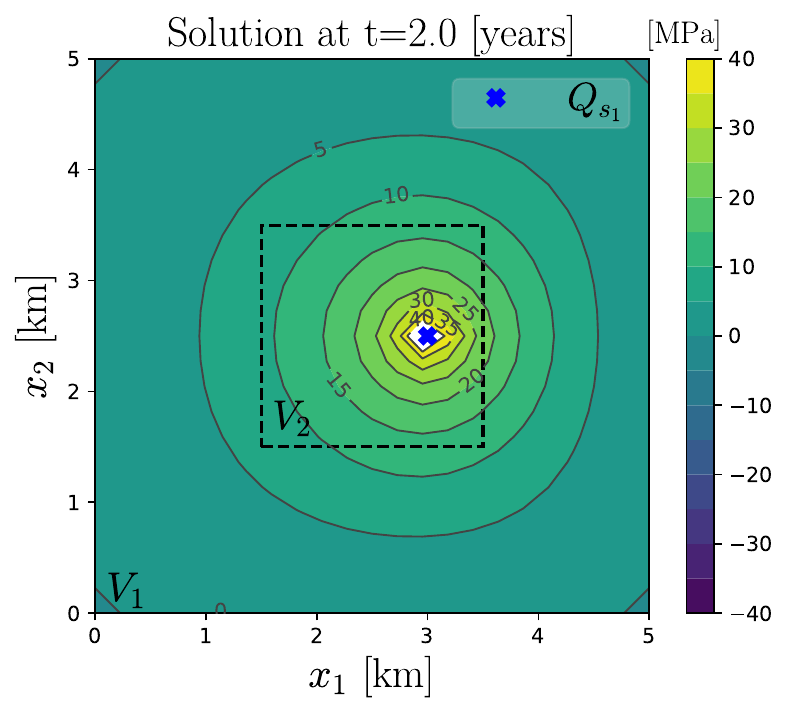}
  \caption{Solution, $u(x,t)$, of the pressure's reservoir at different times, with constant injection rate, $Q_{s_1}=15$ [m$^3$/hr]. No control is applied. The solution presents high-pressure profiles in wide areas next to the injection point. Observing the contour lines, the steady state is reached after two years.}
  \label{fig:u_no}
\end{figure}

Figure \ref{fig:u_no} illustrates the evolution of pressure across the reservoir at different times, without the constant injection rate. The pressure experiences a gradual rise over extensive areas near the injection point, eventually stabilizing at approximately two years.

In the case of an Enhanced Geothermal System \cite{Cornet2019}, there is an interest in increasing the permeability between two wells by creating a small network of cracks to facilitate fluid circulation between them \cite{https://doi.org/10.1002/nag.2330}. This creation of cracks would result in localized microseismicity in the region surrounding the wells. 

Therefore, the control problem addressed in this work aims to achieve a controlled increase in the SR in a small region surrounding certain wells (\textit{e.g.}, in region $V_2$, as depicted in Figure \ref{fig:reservoir_no}), while ensuring that the SR remains constant and equal to one over the larger area of the reservoir (\textit{e.g.}, in region $V_1$, as depicted in Figure \ref{fig:reservoir_no}).

In other words, the objective of this work is to design the control input $\bar{Q}_c$ driving the output $y \in [C^0(T)]^{m_c}$ defined as
\begin{equation}
\begin{split}
  y &= [h_1,...,h_{m_c}]^T, \\
  h_i &= \ln(R_i), \quad i = 1,...,m_c,
\end{split}
  \label{eq:output}
\end{equation}
of the underlying BVP \eqref{eq:diff}--\eqref{eq:SR} to desired references $r(t) \in [C^0(T)]^{m_c}$, $r(t)=[r_1(t),...,r_{m_c}(t)]^T$. This process is known as tracking. Note that \( h_i(t) \) is not directly the average SR. However, if we make $y(t)$ approach to $r(t)$, the original average SR outputs, $R_i(t)$, will tend toward the desired reference $\bar{r}_i(t) = e^{r_i(t)}$ for $i = 1,...,m_c$, thus achieving the goal of controlling indirectly the average SR.

Furthermore, an additional number of flux restrictions, $m_r \in \mathbb{N}$, for the fluid injections, $\bar{Q}_c$ will be considered. We will impose the weighted sum of the injection rates, $\bar{Q}_c(t)$ to be equal to a time-variant, possibly intermittent production rate as
\begin{equation}
    W \bar{Q}_c(t) = D(t),
    \label{eq:restriction}
\end{equation}
where $W \in \mathbb{R}^{m_r \times (m_c+m_r)}$ is a full rank matrix whose elements represent the weighted participation of the well's fluxes for ensuring the demand $D(t) \in [L^\infty(T)]^{m_r}$. Furthermore, the number of inputs of system \eqref{eq:diff}, $m=m_c+m_r$, is equal to the sum of required SR to be controlled, $m_c$, and the number of flux restrictions over the injections points, $m_r$\cite{b:Gutierrez-Stefanou-2024}.
This flux restriction is essential for realistic injection and extraction plans in underground reservoirs, as it reflects the need to meet  different types of energy requirements and storage.

\begin{figure}[ht!]
  \centering 
  \includegraphics[width=0.6\textwidth,keepaspectratio]{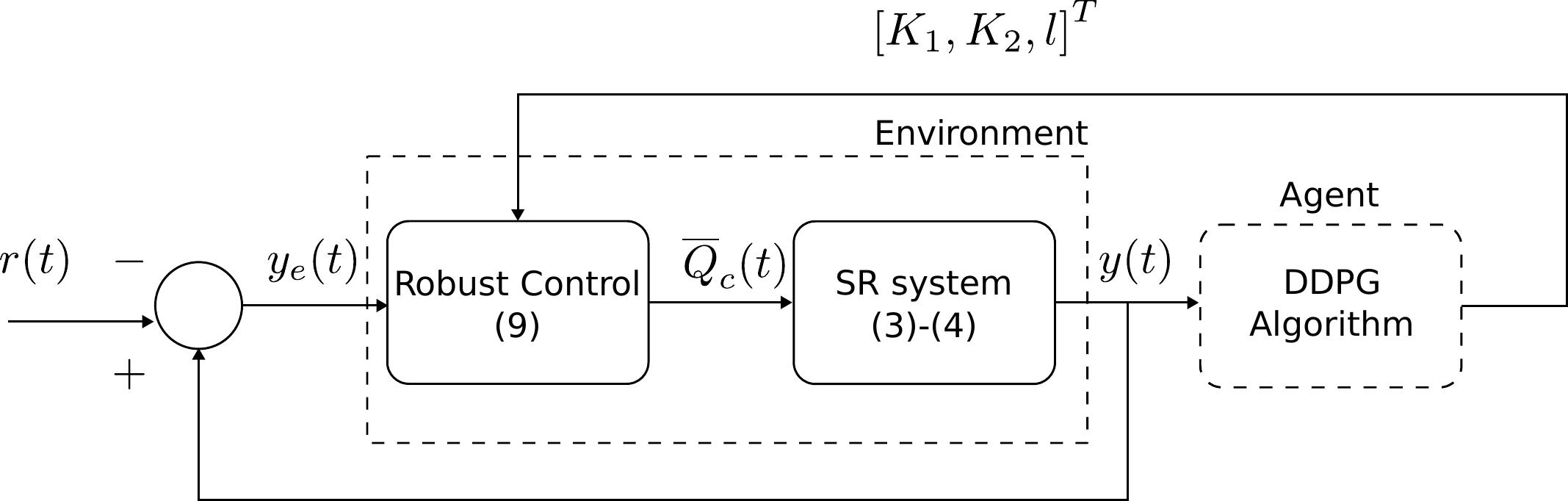}
  \caption{Schematic representation of the control-RL strategy for minimizing induced seismicity over an underground reservoir.}
  \label{fig:diagram}
\end{figure}

In summary, solving the tracking problem over the output \eqref{eq:output} and imposing the flux restriction \eqref{eq:restriction} on the injection fluid will minimize the effects of induced seismicity on the underground reservoir while accommodating various types of energy demand and production constraints. Furthermore, to address a more realistic scenario, the exact knowledge of the system parameters in \eqref{eq:diff} and \eqref{eq:SR} will be unknown. For that purpose, we will design a control-RL strategy where a robust control will be used for performing the tracking over the desired SR references while an RL approach will select its gains based on a suitable reward system. A schematic representation of such a strategy is shown in \ref{fig:diagram}. This design will be addressed in the next sections.

\section{ROBUST CONTROL DESIGN}
\label{sec:control}

Following the control design\cite{b:Gutierrez-Stefanou-2024}, let us define an error variable, $y_e \in [C^0(T)]^{m_c}$, as follows
\begin{equation}
 y_e(t) = y(t)-r(t),
\label{eq:error}
\end{equation}
where $y(t)$ is the SR output \eqref{eq:output} and $r(t)$ are the references to be followed. Then, the control $\bar{Q}_c(t)$ is given by
\begin{equation}
\begin{split}
  \bar{Q}_c(t) &= \overline{W} Q_c(t) + W^T (W W^T)^{-1}D(t)\\
  Q_c(t) &= B_0^{-1} \left(-K_1 \Sabs{y_e}^{\frac{1}{1-l}} + \nu + \dot{r}\right), \\
  \dot{\nu} &= -K_2 \Sabs{y_e}^{\frac{1+l}{1-l}},
\end{split}
\label{eq:Qsr}
\end{equation}
where $\overline{W} \in \mathbb{R}^{(m_c+m_r)\times m_c}$ is the null space of the weight matrix $W$, and $K_1,K_2 \in \mathbb{R}^{m_c \times m_c}$ are matrices to be designed. The matrix $B_0 = [b_{ij}]\in \mathbb{R}^{m_c \times m_c}$ is a nominal matrix defined according to the points of injection, $(x_c^1,...,x_c^{m})$, and the regions, $V_i$, as
\begin{equation}
\begin{split}
  b_{ij} &= \left\{ \begin{array}{c}
  \frac{f_0}{t_{a_0} \dot{\tau}_{0_0} \beta_0 V_{i_0}} \\ 
  0
  \end{array} 
  \begin{array}{c}
  \textup{if} \quad x_c^j \in V_i \\ 
  \textup{if} \quad x_c^j \notin V_i
  \end{array} \right. ,
  \begin{array}{l}
  i \in [1,m_c] \\ 
  j \in [1,m_c]
  \end{array} ,
\end{split}
  \label{eq:B0}
\end{equation}
where the subscript `0' corresponds to the nominal values of the system's parameters (see \cite{b:Gutierrez-Stefanou-2024} for more details).

The second and third equation of \eqref{eq:Qsr} are known as a Multi-Input-Multi-Output Super-Twisting controller \cite{b:9901971,b:Mathey-Moreno-2022}. It is composed by a static part, $Q_c(t)\;\in [C^0(T)]^{m_c}$, and a dynamic part, $\nu(t)\;\in [C^0(T)]^{m_c}$, which is an integral extension. This control depends on the freely chosen parameter $l \in [-1,0]$ and exhibits varying characteristics depending on the value of $l$. When $l=-1$, it features a discontinuous integral term, while for $l \in (-1,0]$, the control function is continuous, degenerating to a linear integral control when $l=0$.

Notably, the controller is designed with minimal information about the system \eqref{eq:diff}--\eqref{eq:SR}, requiring only the measurement of the output $y(t)$ and the knowledge of the nominal matrix $B_0$. Note also that if we replace the first equation of \eqref{eq:Qsr} in \eqref{eq:restriction}, the demand over the controlled injection points will be strictly fulfilled at any time $t$. The tracking result for the output \eqref{eq:SR}--\eqref{eq:output} is then in force.

\textit{Let system \eqref{eq:diff}--\eqref{eq:SR} be driven by the control \eqref{eq:Qsr} with some $K_1>0$, $K_2>0$ and $B_0$. Then, the error variable \eqref{eq:error} will tend to zero in finite-time if $l=[-1,0)$, or exponentially if $l=0$.} In other words, it is theoretically possible to adjust the fluid flux of the wells in an underground reservoir and achieve the desired control objectives in terms of the SR, while achieving production constraints. (See \cite{b:Gutierrez-Stefanou-2024} for the mathematical derivation of the proof and further details of the control algorithm.)

In principle, it is necessary to have some bounds on the uncertainties and perturbations of system \eqref{eq:diff}--\eqref{eq:SR} for the selection of the gains $K_1$, $K_2$ and $l$, to ensure the tracking of the output \eqref{eq:output}. However, obtaining such bounds is extremely challenging in a realistic underground reservoir where exact measurements of system parameters may not be feasible, or where there may be unmodeled dynamics. To address this issue, we will employ a RL algorithm to select these gains in real-time, based on the maximization of a reward system. 

\section{DEEP REINFORCEMENT LEARNING ALGORITHM}
\label{sec:rl}

RL allows the learner (\textit{i.e.} software agent) to determine an optimal behaviour inside an environment (\textit{i.e.}, the set of all states, actions and rewards the agent can take) that will provide the maximum cumulative reward (\textit{i.e.}, a feedback signal from the environment reflecting how well the agent is performing).

Central to these methods is the notion of reward, which the algorithm tries to maximize by transitioning to different states of the environment. The transitions between the different states of the environment take place according to the policy followed by the agent and the underlying model. The policy is a mapping between the states of the environment and the actions available to the agent. For these methods to work, the response of the environment to each action taken by the agent needs to be known, \textit{i.e.}, full knowledge of the underlying model is needed. This can be challenging in systems with large state spaces and continuous actions (the gains $K_1,K_2,l$ of the control \eqref{eq:Qsr}) such as the underground reservoir system.

To account for this, model-free approaches are more suitable \cite{bucsoniu2018reinforcement}. In the model-free framework, the agent learns using estimates of the accumulated reward, in a process called value iteration \cite{bertsekas2019reinforcement,sutton2018reinforcement}. A deep neural network can then be used to interpolate the reward estimates among adjacent states.

To allow the agent to explore the state space of the optimisation problem such that better estimates of the accumulated reward can be drawn, a policy gradient algorithm is used starting from a random policy, that is progressively improved through gradient updates \cite{sutton1999policy,williams1992simple,silver2014deterministic}.

The two methods of value iteration and policy gradients can be combined in the so-called actor-critic algorithms of RL \cite{geron2022hands,grondman2012survey,sutton2018reinforcement}. This allows for an efficient exploration of the state space of the problem and better estimates of the accumulated reward. Under this context, we call an ``actor'' the part of the agent that is responsible for selecting actions based on the current policy, and we call a ``critic'' a deep neural network that learns to predict the action values. In contrast to the actor, the critic learns an approximation of the accumulated reward over the state-action space. This is done using appropriate interpolation weights. The critic then provides to the actor the action value associated with the actor's action, which is an approximation of the accumulated reward.



In essence, the critic gains knowledge of the states and the rewards of the task, during the evaluation of the actor policy and predicts the estimated accumulated reward for each given state. Then, the actor uses the critic's estimates of the reward to update its policy. This speeds up the policy evaluation step of the actor since it no longer needs the episode to finish before it starts updating the weights.   

Inspired by \cite{b:Timos_Stefanou_2021}, the actor-critic algorithm known as the Deep Deterministic Policy Gradient (DDPG) algorithm \cite{b:lillicrap2019continuous} has been chosen. The agent learns to meet the objectives of interacting with the reservoir by changing the gains $K_1$, $K_2$, and $l$ of the control \eqref{eq:Qsr} during the reservoir's exploitation. This way, the agent ensures that induced seismicity is mitigated and the energy demands are met.

The environment where the DDPG algorithm will be trained is represented by the feedback connection of the simplified model of an average SR system over a 3D underground reservoir, governed by equations \eqref{eq:diff}--\eqref{eq:SR}, and the robust control of equation \eqref{eq:Qsr}. The agent consists of the actor-critic network of the DDPG algorithm and takes only the tracking output, $y(t)$, as observation for the calculation of the gains. The gains range $K_1 \in [0, \, 5 \times 10^{-4}] \, \mathbb{I}$, $K_2 \in [0, \, 5 \times 10^{-4}] \, \mathbb{I}$ and $l \in [-1, \, 0]$ of the control \eqref{eq:Qsr} are considered to be the action of the RL algorithm.

As stated before, the control \eqref{eq:Qsr} has been tested at minimizing induced seismicity in the underground reservoir\cite{b:Gutierrez-Stefanou-2024}. Nevertheless, to optimize such control to account for other performance targets, we define a normalized reward based on the error \eqref{eq:error} and the control \eqref{eq:Qsr}, as
    \begin{equation}
    \begin{split}
        Reward = \frac{1}{n}\left[(1-\alpha)e^{-y_{ref}\norm{y_e(t)}} + \alpha e^{-Q_{ref}\norm{Q_c(t)}} \right],
    \end{split}
    \label{eq:reward}
    \end{equation}
where $n$ is the total number of steps per episode, and $\alpha \in [0, 1]$ is a hyperparameter controlling the trade-off between minimizing the tracking error and minimizing the norm of the control signal. The constants $y_{ref}=1$ [-] and $Q_{ref}=1 \times 10^6$ [m$^3$/hr] are used to pass dimensionless quantities in the exponential functions. In this formulation, both terms of \eqref{eq:reward} reach their maximum value when the norms are minimized. The hyperparameter $\alpha$ balances between these two objectives. The choice of this reward system aims to achieve optimal precision in the tracking error, $y_e(t)$, while minimizing the control power of the wells.

Table \ref{tab:hyper} shows the selection of hyperparameters of the DDPG algorithm for its training. The results will be shown in the next Section.

\begin{table}
    \begin{subtable}{.5\linewidth}
    \centering
    \begin{tabular}{ccc}
      \toprule
      \textbf{Parameter} & \textbf{Description} & \textbf{Value} \\
      \midrule
      $\tau$ & Update rate & $0.005$ \\
      $\gamma$ & Discount factor & $0.99$ \\
      $\sigma$ & Standard deviation of noise & $0.1$ \\
      $\alpha$ & Reward parameter & $0.5$ \\
      \bottomrule
    \end{tabular}
    \end{subtable}%
    \begin{subtable}{.5\linewidth}
    \begin{tabular}{ccc}
      \toprule
      \textbf{} & \textbf{Actor Network} & \textbf{Critic Network} \\
      \midrule
      First layer & 400 Neurons & 400 neurons \\
       & ReLU activation function & ReLU activation function \\
      Second layer & 300 Neurons & 300 neurons \\
       & ReLU activation function & ReLU activation function \\
      Output layer & Linear activation function & Sigmoid activation function \\
      Learning rate & $0.001$ & $0.0001$ \\
      \bottomrule
    \end{tabular}
    \end{subtable}    
    \caption{Selected hyperparameters and network architecture of the DDPG.}
    \label{tab:hyper}
\end{table}

\section{SIMULATIONS AND DISCUSSION}
\label{sec:sim}

\begin{figure}[ht!]
  \centering 
  \includegraphics[width=0.45\textwidth,keepaspectratio]{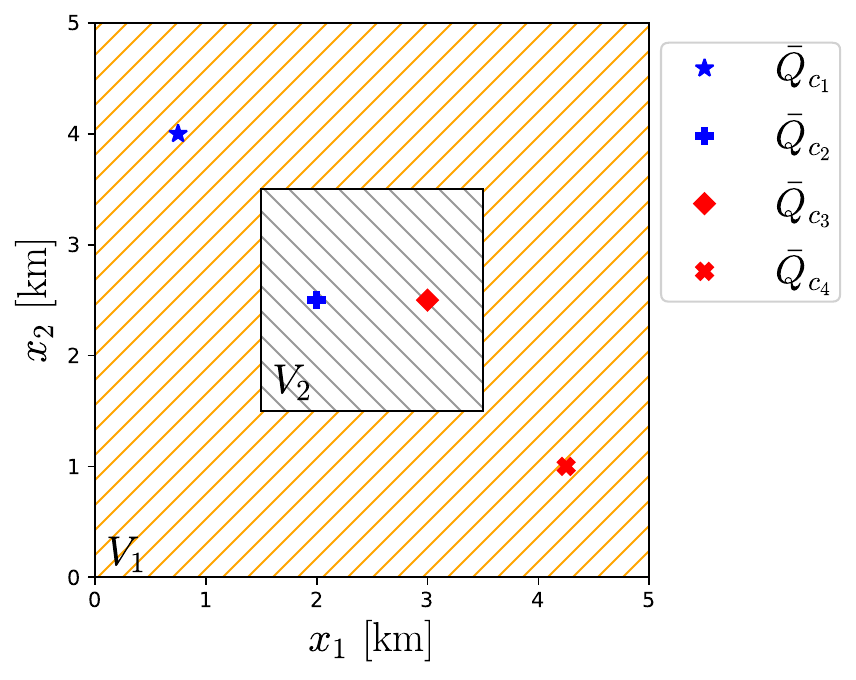}
  \caption{Regions $V_1$ and $V_2$ and location of the injection wells.}
  \label{fig:reservoir_control}
\end{figure}

To demonstrate our control-RL approach, numerical simulations of \eqref{eq:diff} and \eqref{eq:SR} have been done in Python using the same parameters and numerical methods performed in section \ref{sec:problem}. Following the same example, we consider two different regions, $V_1$,$V_2$ over which we calculate the SR, \textit{i.e.}, $y(t)=[h_1(t),h_2(t)]^T$, $m_c=2$. We will apply two flux restrictions over the fluid injections, \textit{i.e.}, $m_r=2$. This results in having a total of four injection points to be needed ($\bar{Q}_c(t)=[\bar{Q}_{c_1}(t),\bar{Q}_{c_{2}}(t),\bar{Q}_{c_{3}}(t),\bar{Q}_{c_{4}}(t)]^T$), whose location is depicted in Fig. \ref{fig:reservoir_control}. The initial conditions of the systems \eqref{eq:diff} and \eqref{eq:SR} were set as $h_1(0)=h_2(0)=0$ (consequently, $R_1=R_2=1$) and $u(x,0)$ was chosen as a random number between $[-10,10]$ [kPa].

\begin{figure}[ht!]
     \centering
     \begin{subfigure}[b]{0.45\textwidth}
         \includegraphics[width=\textwidth,keepaspectratio]{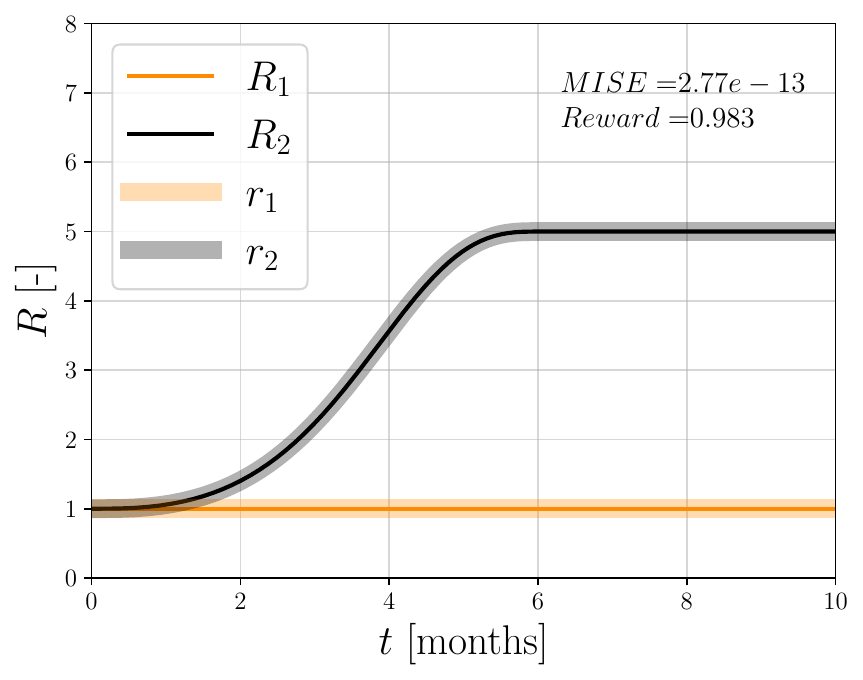}
         \caption{Seismicity rate in regions $V_1,V_2$ using control-RL approach.}
     \end{subfigure}
     \hspace{15pt}
     \begin{subfigure}[b]{0.45\textwidth}
         \includegraphics[width=\textwidth,keepaspectratio]{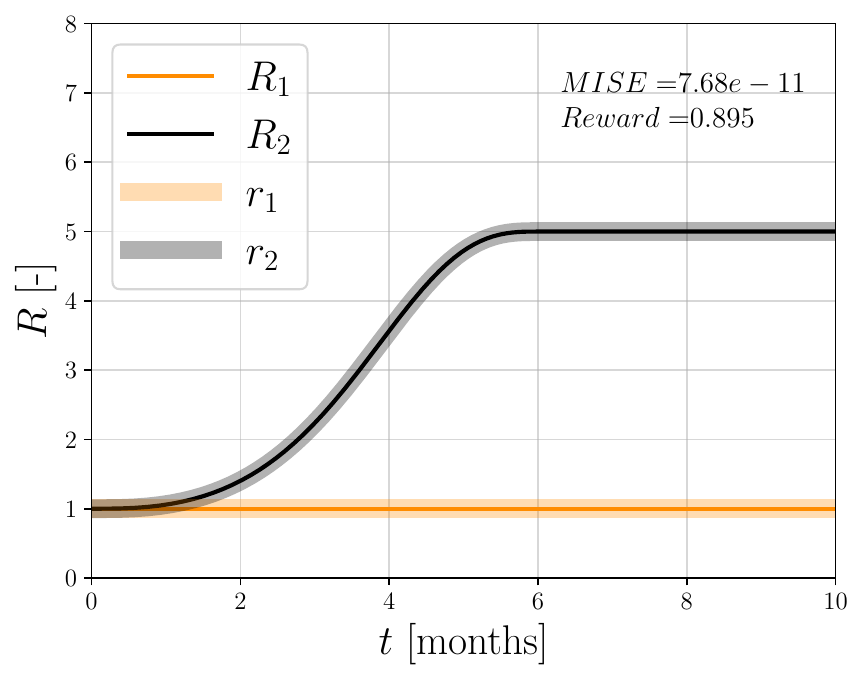}
         \caption{Seismicity rate in regions $V_1,V_2$ using control approach.}
     \end{subfigure}
     \begin{subfigure}[b]{0.45\textwidth}
         \includegraphics[width=\textwidth,keepaspectratio]{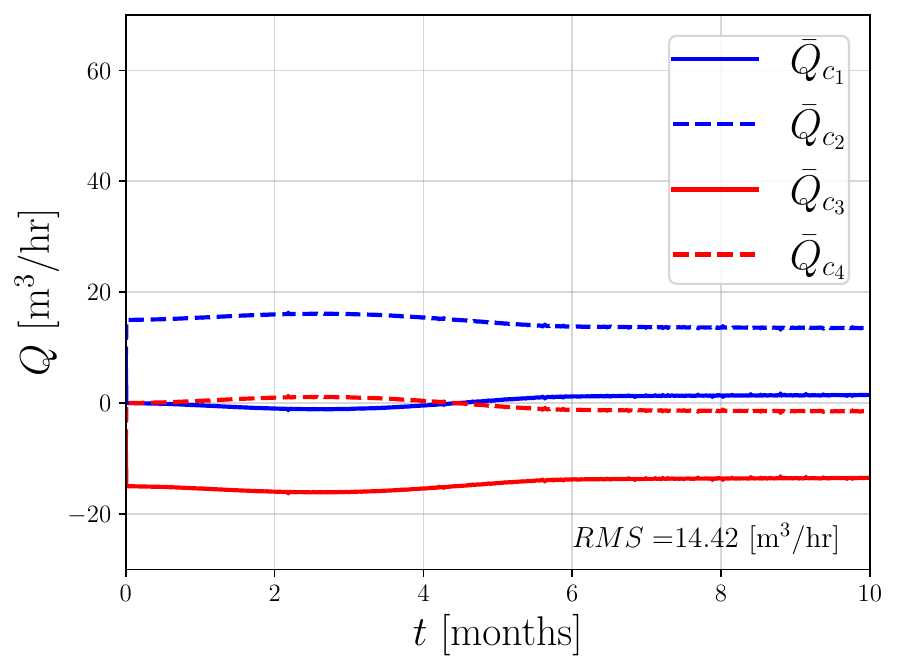}
         \caption{Controlled flux inputs using control-RL approach.}
     \end{subfigure}
     \hspace{15pt}
     \begin{subfigure}[b]{0.45\textwidth}
         \includegraphics[width=\textwidth,keepaspectratio]{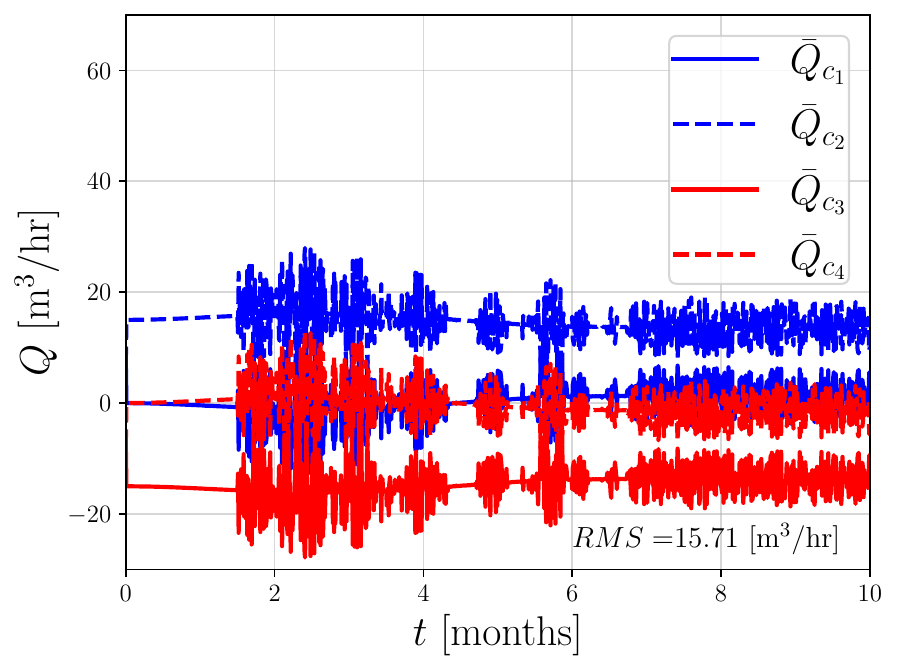}
         \caption{Controlled flux inputs using control approach.}
     \end{subfigure}
    \caption[b]{Seismicity rate and controlled flux inputs $\bar{Q}_{c_1}(t),\bar{Q}_{c_{2}}(t),\bar{Q}_{c_{3}}(t),\bar{Q}_{c_{4}}(t)$ under constant demand.}
  \label{fig:SR1}
\end{figure}

\begin{figure}[ht!]
     \centering
     \begin{subfigure}[b]{0.45\textwidth}
         \includegraphics[width=\textwidth,keepaspectratio]{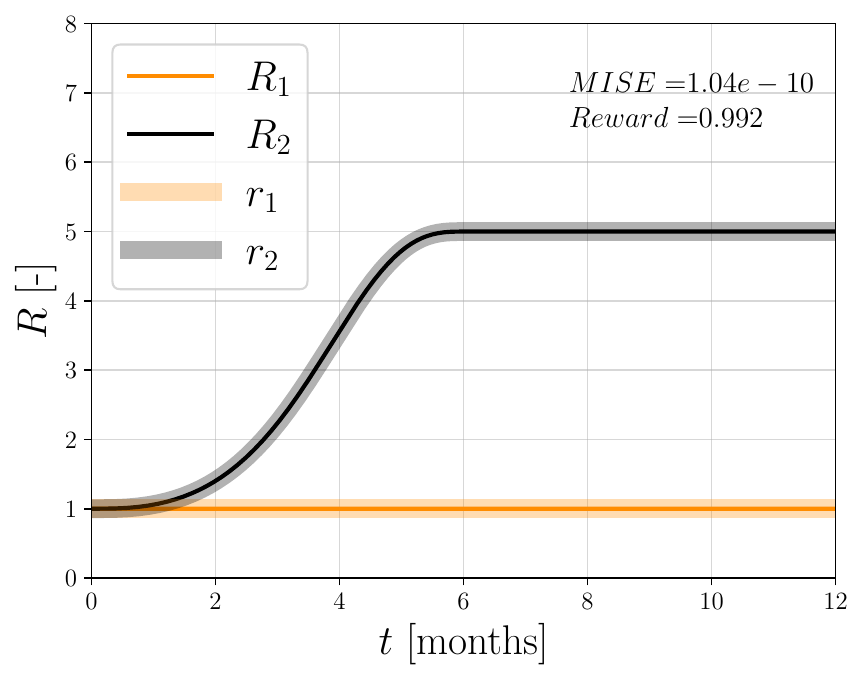}
         \caption{Seismicity rate in regions $V_1,V_2$ using control-RL approach.}
     \end{subfigure}
     \hspace{15pt}
     \begin{subfigure}[b]{0.45\textwidth}
         \includegraphics[width=\textwidth,keepaspectratio]{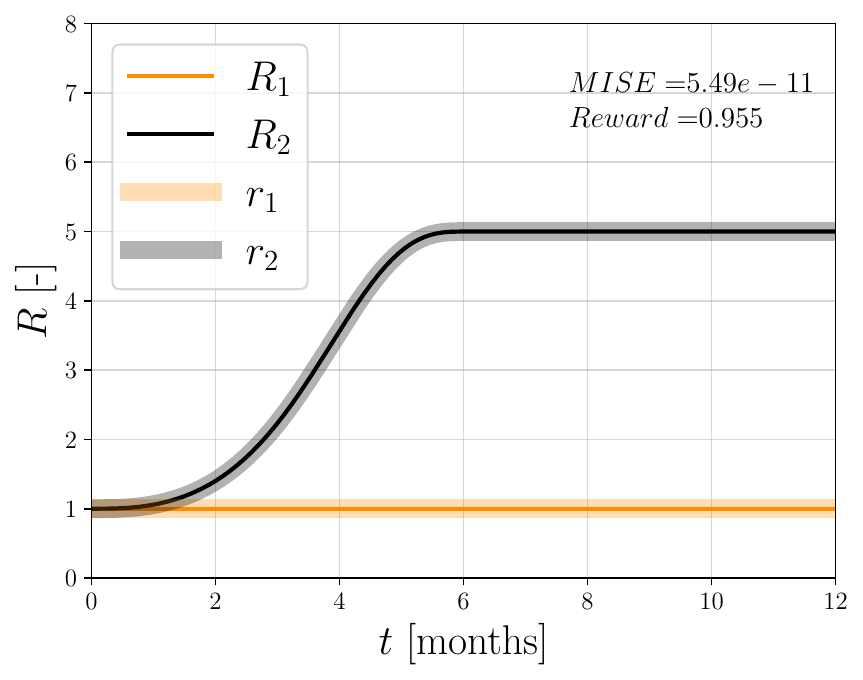}
         \caption{Seismicity rate in regions $V_1,V_2$ using control approach.}
     \end{subfigure}
     \begin{subfigure}[b]{0.45\textwidth}
         \includegraphics[width=\textwidth,keepaspectratio]{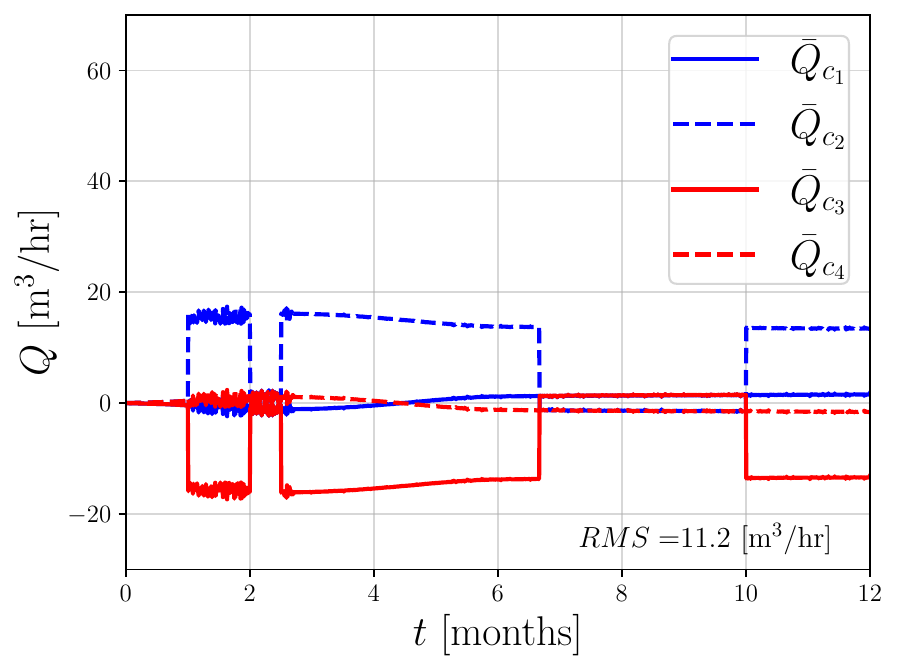}
         \caption{Controlled flux inputs using control-RL approach.}
     \end{subfigure}
     \hspace{15pt}
     \begin{subfigure}[b]{0.45\textwidth}
         \includegraphics[width=\textwidth,keepaspectratio]{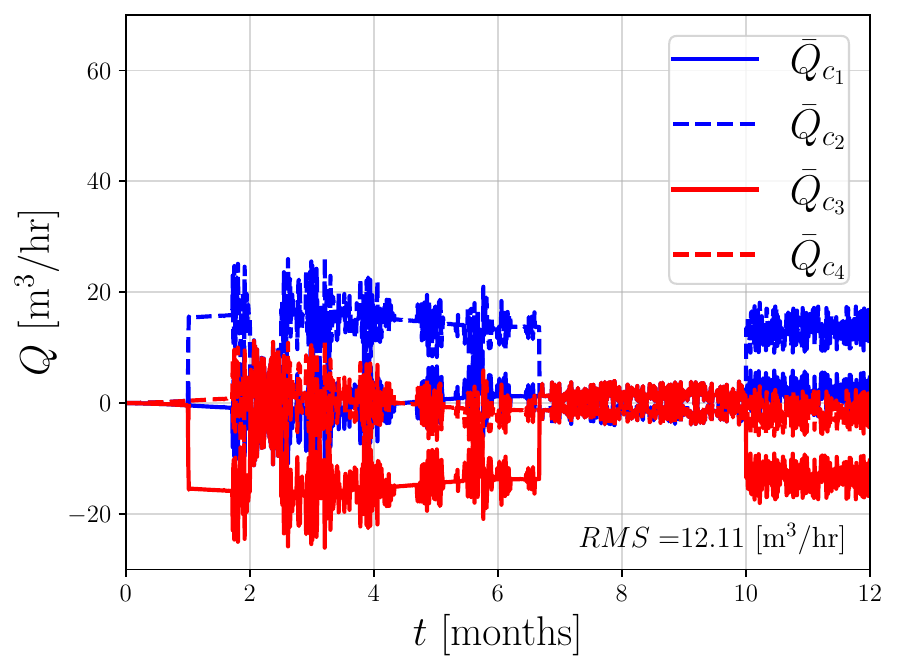}
         \caption{Controlled flux inputs using control approach.}
     \end{subfigure}
    \caption[b]{Controlled flux inputs $\bar{Q}_{c_1}(t),\bar{Q}_{c_{2}}(t),\bar{Q}_{c_{3}}(t),\bar{Q}_{c_{4}}(t)$ under intermittent demand.}
  \label{fig:SR2}
\end{figure}

The reference $r(t)$ was selected as $r(t)=[r_1(t),r_2(t)]^T$, where $r_1(t)=\ln (1)=0$ and $r_2(t)$ is a smooth function that reaches the final value of $\ln(5)$ in $6$ [months] (see the references in Figs. \ref{fig:SR1} and \ref{fig:SR2}, top subfigures). This reference was chosen so that the SR on every region, $V_1,V_2$ converges to the final values of $1$ and $5$, respectively. This selection aims at forcing the SR in the extended region $V_1$ to be the same as the natural one. Regarding, region $V_2$ we opt for an increase of the SR to facilitate the circulation of fluids and thus improve the production of energy, as explained in Section \ref{sec:problem}.

We will apply the two flux restrictions based on the selection of $W$ and $D$ in eq. \eqref{eq:restriction} as
\begin{equation}
\begin{split}
  W &= \left[ 
  \begin{array}{cccc}
  1.01 & 1 & 0 & 0 \\
  0 & 0 & 1 & 1  
  \end{array}
  \right], \quad
  D(t) = \left[ 
  \begin{array}{c}
  Q_s(t) \\
  -Q_s(t) 
  \end{array}
  \right],
\end{split}
  \label{eq:matrices_demand}
\end{equation} 
where $Q_s(t)=15$ [m$^3$/hr]. This selection will induce fluid circulation within the reservoir such that two wells will inject a fluid flux equal to $Q_s(t)$, while the other two wells will extract the same flux from the reservoir. Other scenarios could be considered as well.

The control $\bar{Q}_c(t)$ was designed as \eqref{eq:Qsr} with the nominal matrix $B_0$ selected according to \eqref{eq:B0} as 
\begin{equation}
  B_0 = \frac{f_0}{t_{a_0} \dot{\tau}_{0_0} \beta_0} \left[ 
  \begin{array}{cccc}
  \frac{1}{V_{1_0}} & 0 & 0 & \frac{1}{V_{1_0}} \\
  0 & \frac{1}{V_{2_0}} & \frac{1}{V_{2_0}} & 0
  \end{array}
  \right],
  \label{eq:matrices_control}
\end{equation}
where the subscript `0' corresponds to the nominal values of the system's parameters. We have chosen all the nominal values 10$\%$ higher than the real ones, \textit{e.g.}, $f_0=1.1 f$\cite{b:Gutierrez-Stefanou-2024}. 

The gain parameters of the control \eqref{eq:Qsr}, $K_1$, $K_2$, and $l$ were selected according to the model trained by the DDPG algorithm presented in Section \ref{sec:rl}. To compare this strategy, a control $\bar{Q}_c(t)$ with fixed gains $K_1=5 \times 10^{-4} \, \mathbb{I}_2$, $K_2=5 \times 10^{-4} \, \mathbb{I}_2$ and $l=-1$ (without RL gain selection) will be tested and compared using the mean integrated square error ($MISE=\frac{1}{t_{max}} \int_0^{t_{max}} \norm{y_e(t)}^2 \, dt$) and the average power of the control action ($RMS=\sqrt{\frac{1}{t_{max}} \int_0^{t_{max}} \norm{{Q}_c(t)}^2 \, dt}$). These gains were chosen as the largest possible among the feasible range to obtain the best tracking precision (see the action description in Section \ref{sec:rl}).

\begin{figure}[ht!]
     \centering
     \begin{subfigure}[b]{0.45\textwidth}
         \includegraphics[width=\textwidth,keepaspectratio]{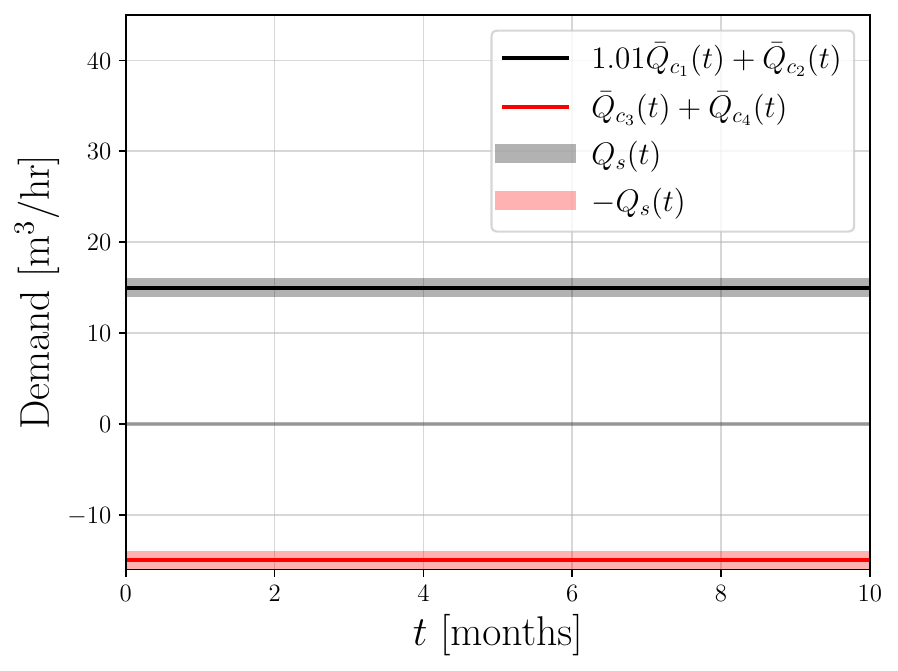}
         \caption{Constant demand on the flux inputs.}
     \end{subfigure}
     \hspace{15pt}
     \begin{subfigure}[b]{0.45\textwidth}
         \includegraphics[width=\textwidth,keepaspectratio]{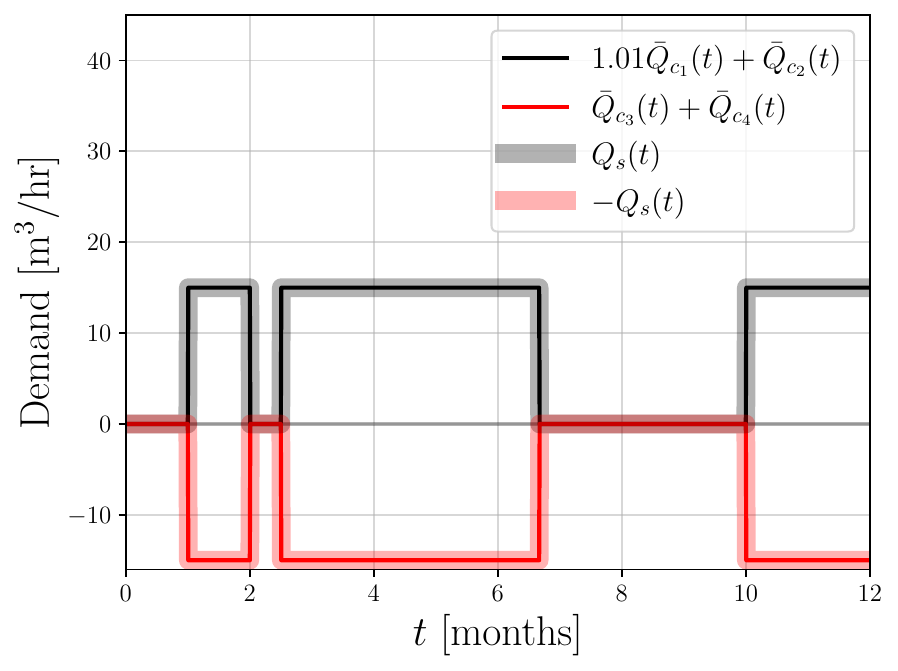}
         \caption{Intermittent demand on the flux inputs.}
     \end{subfigure}
    \caption[b]{Different types of demand, $D(t)$, used in the simulations.}
  \label{fig:demand}
\end{figure}

The results are illustrated in Figure \ref{fig:SR1}. Both approaches successfully prevent induced seismicity by ensuring tracking of the desired seismicity rate while adhering to the specified flux restriction (see Fig. \ref{fig:demand}, left subfigure). However, the control-RL strategy achieves this task with a lower MISE, less energy consumption (RMS value) and higher accumulated reward. This distinction is shown in Fig. \ref{fig:SR1}, bottom subfigures, where the control strategy exhibits more pronounced oscillations in the generated control fluxes than the control-RL method. 

To test a more realistic scenario, a challenging intermittent demand pattern is introduced, as depicted in Fig. \ref{fig:demand} (right side), following a pattern similar to \cite{b://doi.org/10.1029/2019JB019134}. This demand plan presents abrupt variations between the injection flux $Q_s(t)$ and zero. The results are displayed in Fig. \ref{fig:SR2}. It is demonstrated that both strategies effectively achieve the control objectives. However, the control-RL strategy accomplishes this task with lower energy consumption and a higher accumulated reward. 

\begin{figure}[ht!]
     \centering
     \begin{subfigure}[b]{0.45\textwidth}
         \includegraphics[width=\textwidth,keepaspectratio]{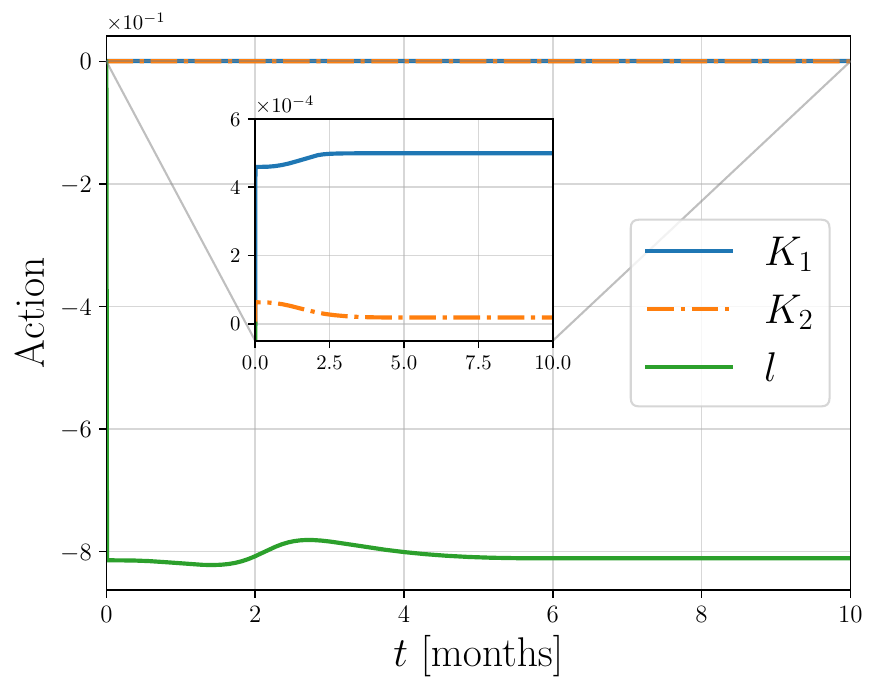}
         \caption{Gains evolution under constant demand on the flux inputs.}
     \end{subfigure}
     \hspace{15pt}
     \begin{subfigure}[b]{0.45\textwidth}
         \includegraphics[width=\textwidth,keepaspectratio]{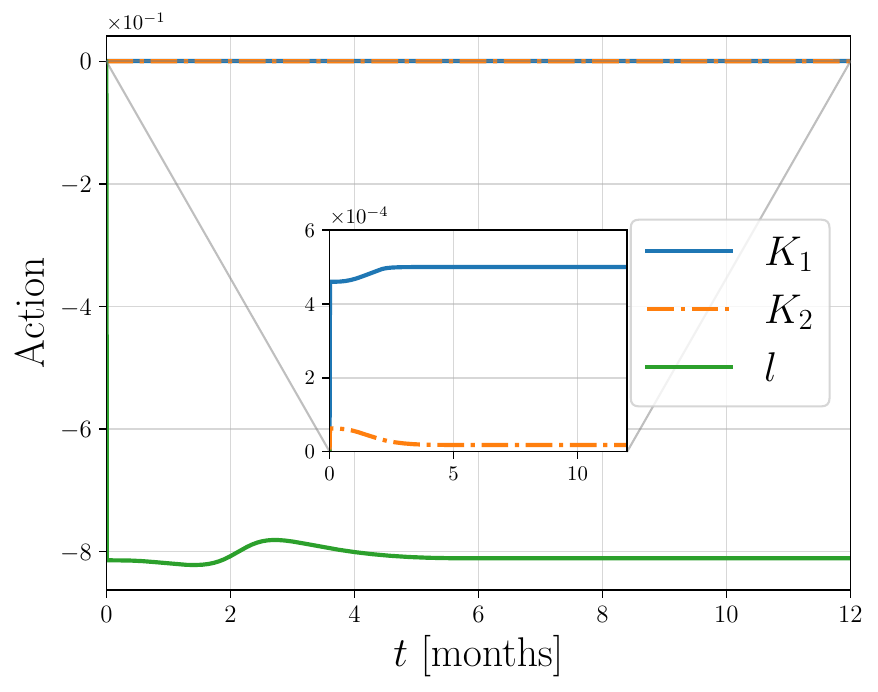}
         \caption{Gains evolution under intermittent demand on the flux inputs.}
     \end{subfigure}
    \caption[b]{Evolution of the gains, $K1,K2,l$, thanks to the control-RL approach.}
  \label{fig:gains}
\end{figure}

Figure \ref{fig:gains} shows how the gains are evolving during both cases to achieve these tasks. One can notice that the trained RL model chooses a high gain as $K_1$, a low gain as $K_2$, and a homogeneous control between the linear and discontinuous control ($l \approx -0.8$).

\begin{figure}[ht!]
     \centering
     \begin{subfigure}[b]{0.4\textwidth}
         \includegraphics[width=\textwidth,keepaspectratio]{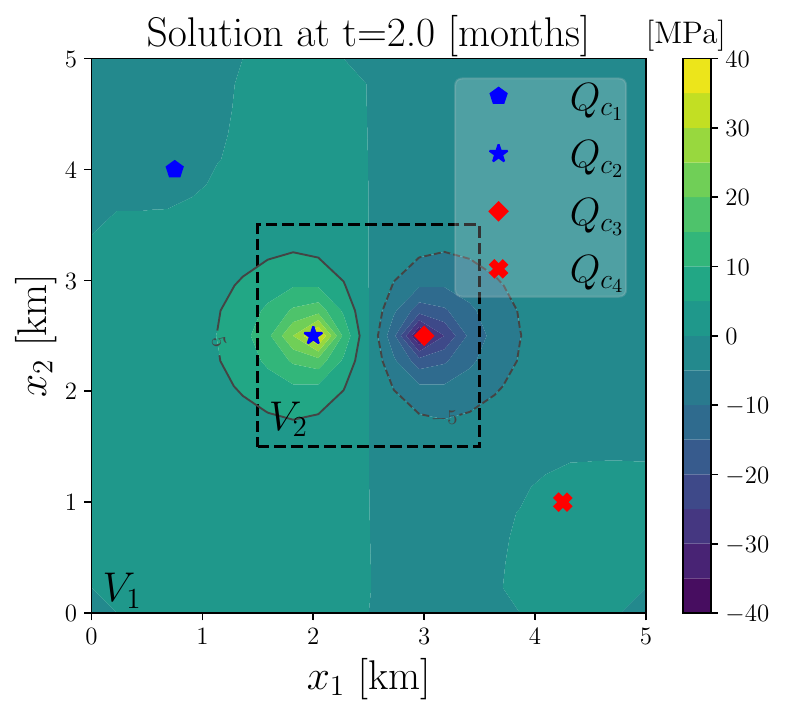}
     \end{subfigure}
     \hspace{15pt}
     \begin{subfigure}[b]{0.4\textwidth}
         \includegraphics[width=\textwidth,keepaspectratio]{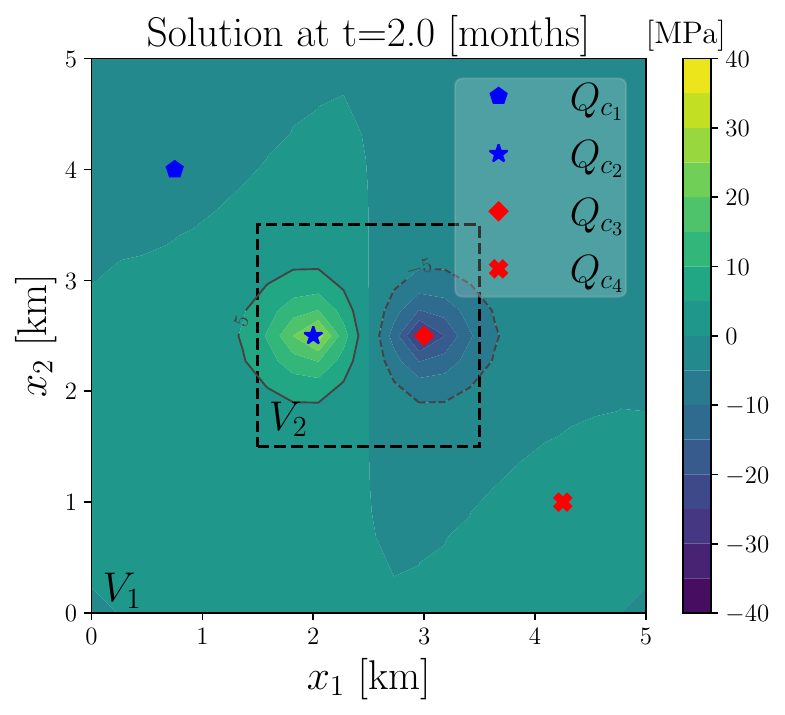}
     \end{subfigure}
     \begin{subfigure}[b]{0.4\textwidth}
         \includegraphics[width=\textwidth,keepaspectratio]{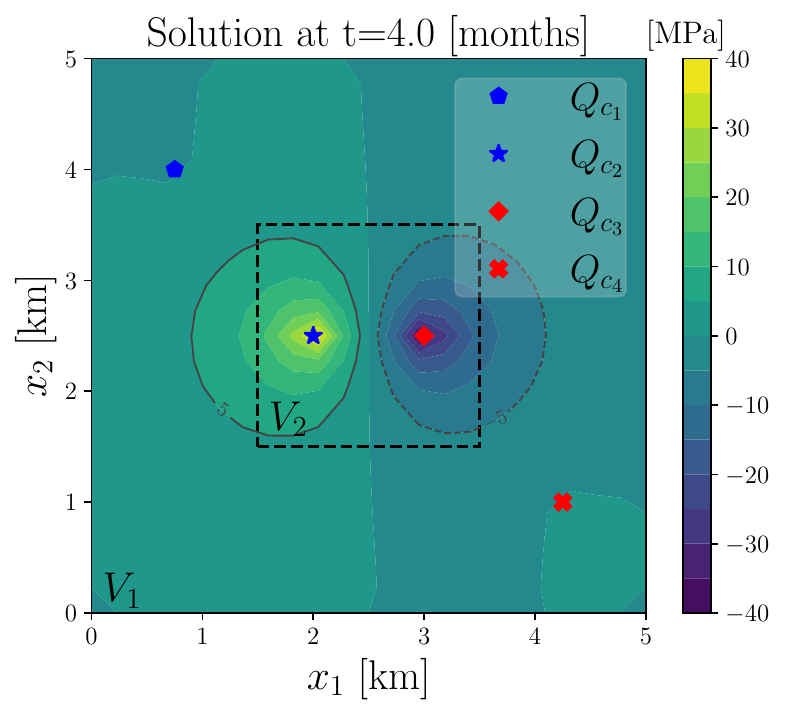}
     \end{subfigure}
     \hspace{15pt}
     \begin{subfigure}[b]{0.4\textwidth}
         \includegraphics[width=\textwidth,keepaspectratio]{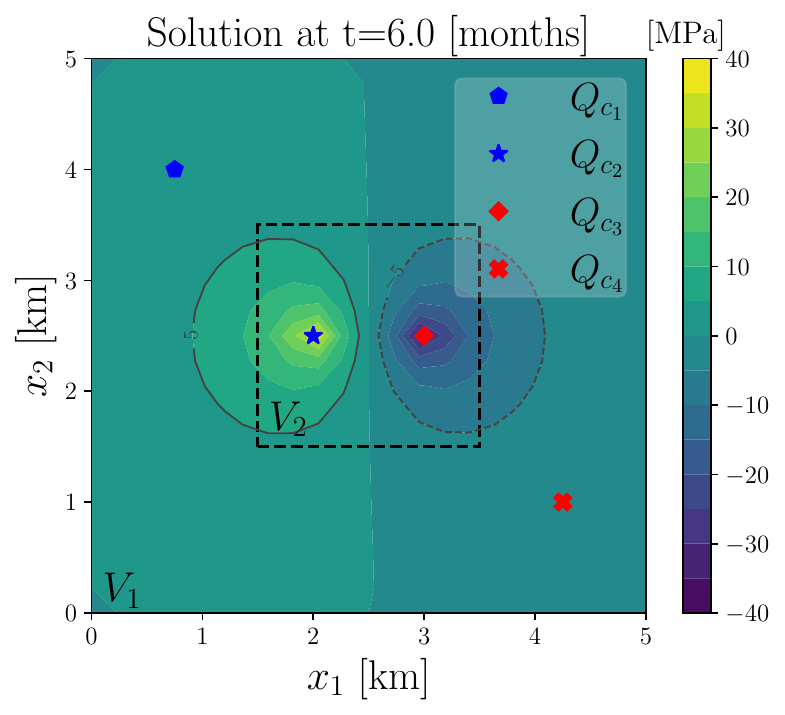}
     \end{subfigure}
     \begin{subfigure}[b]{0.4\textwidth}
         \includegraphics[width=\textwidth,keepaspectratio]{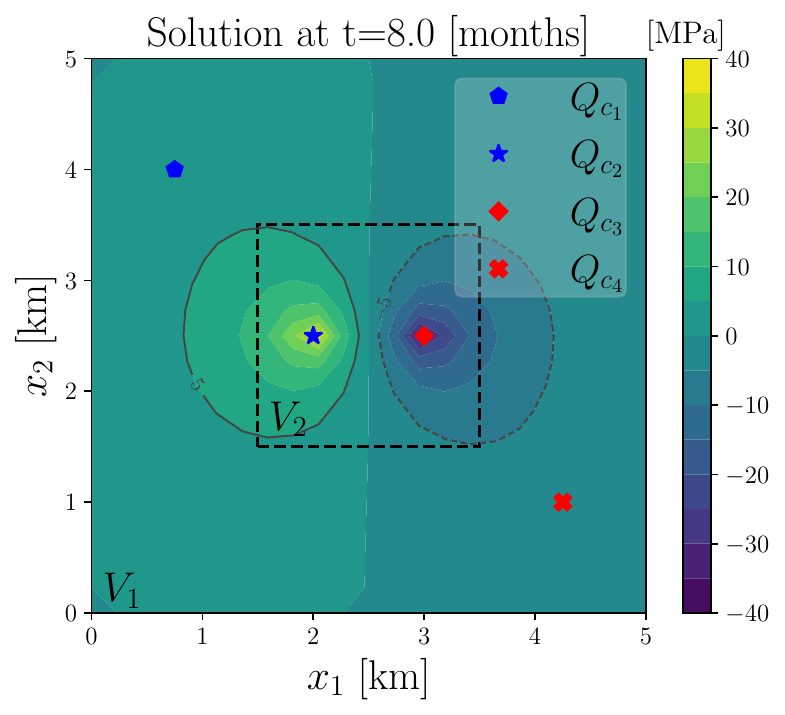}
         \caption{Pressure distribution using control-RL approach and constant demand.}
     \end{subfigure}
     \hspace{15pt}
     \begin{subfigure}[b]{0.4\textwidth}
         \includegraphics[width=\textwidth,keepaspectratio]{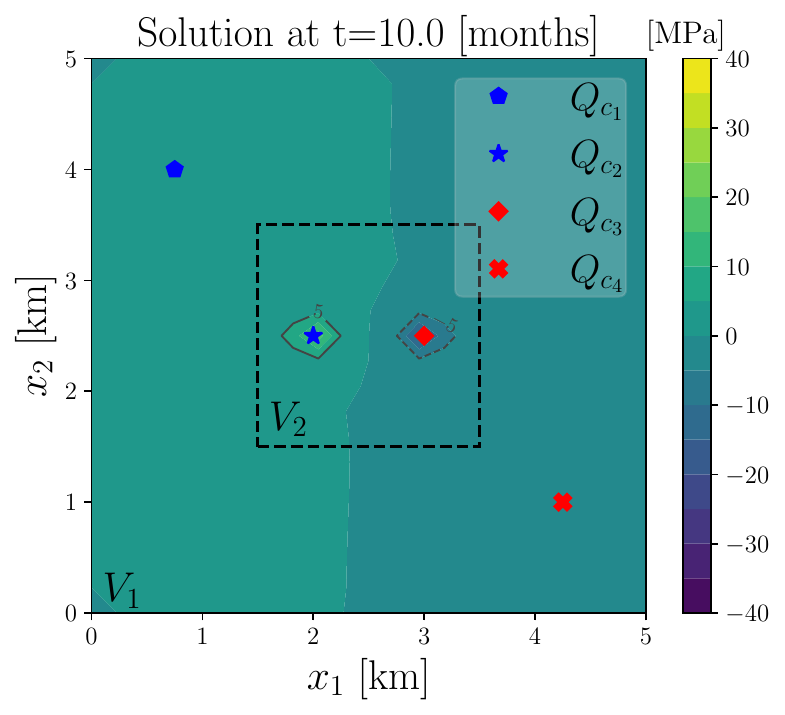}
         \caption{Pressure distribution using control-RL approach and intermittent demand.}
     \end{subfigure}
    \caption[b]{Fluid pressure distribution, $u(x,t)$, in the reservoir at different times. The control-RL strategy prevents the propagation of high-pressure profiles throughout the underground reservoir in contrast with the induced seismicity example of Fig. \ref{fig:u_no}.}
  \label{fig:u}
\end{figure}

Fig. \ref{fig:u} illustrates the pressure profile $u(x,t)$ at various time points under both demand scenarios. In contrast to the scenario without control (refer to Figure \ref{fig:u_no}), the presented combined strategy (control-RL) successfully prevents the propagation of high-pressure profiles around the underground reservoir, confining the highest pressures around the injection points only.

It is worth noting how the control strategy can address the SR tracking problem by itself\cite{b:Gutierrez-Stefanou-2024}. Yet, the RL approach introduces an additional optimization objective: minimizing the energy consumption of the actuators. This dual focus not only ensures precise SR tracking but also enhances overall system efficiency by reducing energy expenditure and balancing both performance and energy usage according to the reward function. 

One may consider a robust Uncertainty Quantification (UQ) framework (\textit{e.g.}, \cite{b:10.1785/0220230179}) to estimate the system’s uncertainties and perturbations more accurately, thereby avoiding overcompensation with high-gain control and further reducing actuator energy requirements. However, this is beyond the scope of the present work.

Assessing earthquake risk based solely on seismicity rate may present limitations, as earthquake magnitude is often more critical than seismicity rate (\textit{cf.}, Pohang EGS project \cite{KIM2022105098,https://doi.org/10.1029/2019JB018368} vs. Basel EGS project \cite{HARING2008469}). The frequency-magnitude relationship of expected earthquakes can be derived from modified Gutenberg-Richter distributions, as shown in \cite{https://doi.org/10.1002/jgrb.50264} and related works. Additionally, the maximum magnitude of anticipated earthquakes may correlate with the size of activated regions, \( V_i \). The incorporation of such statistical analyses, however, also exceeds the purpose of this work.

Moreover, the inclusion of fault discontinuities warrants further study in real scenarios. Controlling multiple faults within a reservoir (\textit{cf.}, \cite{b:Boyet_2023}) poses challenges due to the complexities and faster spatio-temporal scales of poroelastodynamic phenomena triggered by intermittent injections. Managing this dynamic alongside point-wise SR, as opposed to region-wise constraints, nonlinear flux restrictions, and measurement errors, among other factors, remains an ongoing area of investigation. Some first steps towards this direction are presented in \cite{b:Gutierrez-Stefanou-Plestan-2024}.

\section{CONCLUSIONS}
\label{sec:conclusions}

The paper presents an integrating control theory and reinforcement learning strategy to mitigate induced seismicity while maintaining fluid circulation for energy production in underground reservoirs. The robust control mechanism leverages region-based seismicity rate averages to track desired seismicity rates across diverse regions of underground reservoirs. Given the inherent uncertainties in system parameters and potential errors in sensor measurements, the reinforcement learning algorithm optimizes control gains to minimize tracking errors while optimising the energy consumption of the actuators.

Numerical simulations demonstrate the efficacy of the proposed methodology using a simplified underground reservoir model. This new approach opens a direction for future research for using artificial intelligence to address more optimization objectives, but also, to account for more intricate and realistic phenomena, including poroelastodynamic effects, discrete-time dynamics, optimization with nonlinear constraints on control well fluxes, and handling multiple faults.

\bmsection*{Author contributions}

\textbf{Diego Guti\'errez-Oribio:} Methodology, Software, Writing –
original draft, Validation, Visualization. \textbf{Alexandros Stathas: } Writing – review \& editing, Software, Validation. \textbf{Ioannis Stefanou:} Conceptualization, Methodology, Software, Writing – review \& editing, Supervision, Funding acquisition.

\bmsection*{Acknowledgments}

The authors want to acknowledge the European Research Council's (ERC) support under the European Union’s Horizon 2020 research and innovation program (Grant agreement no. 101087771 INJECT) and the Region Pays de la Loire and Nantes M\'etropole under the Connect Talent programme (CEEV: Controlling Extreme EVents - Blast: Blas LoAds on STructures). The authors would like to thank Dr. E. Papachristos for their fruitful discussions.

\bmsection*{Conflict of interest}

The authors declare no potential conflict of interest.

\bmsection*{Data availability statement}

The data that support the findings of this study are available from the corresponding author upon reasonable request.

\bibliography{Bibliografias}




\end{document}